\documentclass[journal=jacsat,manuscript=article]{achemso}

\usepackage[version=3]{mhchem} 
\usepackage{subcaption}
\usepackage[table]{xcolor}
\usepackage{hyperref}
\usepackage[toc,page]{appendix}
\usepackage{svg}

\author{Peter Myung-Won Pak}
\email{ppak@cmu.edu}
\affiliation{
  Department of Mechanical Engineering, Carnegie Mellon University, Pittsburgh,
  PA, USA
}

\author{Francis Ogoke}
\email{oogoke@andrew.cmu.edu}
\affiliation{
  Department of Mechanical Engineering, Carnegie Mellon University, Pittsburgh,
  PA, USA
}
\author{Andrew Polonsky}
\email{apolon@sandia.gov}
\affiliation{
  Sandia National Laboratories
}
\author{Anthony Garland}
\email{agarlan@sandia.gov}
\affiliation{
  Sandia National Laboratories
}
\author{Dan S. Bolintineanu}
\email{dsbolin@sandia.gov}
\affiliation{
  Sandia National Laboratories
}
\author{Dan R. Moser}
\email{dmoser@sandia.gov}
\affiliation{
  Sandia National Laboratories
}
\author{Michael J. Heiden}
\email{mheiden@sandia.gov}
\affiliation{
  Sandia National Laboratories
}
\author{Amir Barati Farimani}
\email{barati@cmu.edu}
\affiliation{
  Department of Mechanical Engineering, Carnegie Mellon University, Pittsburgh,
  PA, USA
}
\alsoaffiliation{
  Machine Learning Department, Carnegie Mellon University, Pittsburgh, PA, USA
}
\title{
  ThermoPore: Predicting Part Porosity Based on Thermal Images Using Deep
  Learning
}

\abbreviations{IR,NMR,UV}

\begin{document}

\begin{abstract}
  We present a deep learning approach for quantifying and localizing
  \textit{ex-situ} porosity within Laser Powder Bed Fusion fabricated samples
  utilizing \textit{in-situ} thermal image monitoring data. Our goal is to build
  the real time porosity map of parts based on thermal images acquired during
  the build. The quantification task builds upon the established Convolutional
  Neural Network model architecture to predict pore count and the localization
  task leverages the spatial and temporal attention mechanisms of the novel
  Video Vision Transformer model to indicate areas of expected porosity. Our
  model for porosity quantification achieved a $\text{R}^2$ score of 0.57 and
  our model for porosity localization produced an average IoU score of 0.32 and
  a maximum of 1.0. This work is setting the foundations of part porosity
  "Digital Twins" based on additive manufacturing monitoring data and can be
  applied downstream to reduce time-intensive post-inspection and testing
  activities during part qualification and certification. In addition, we seek
  to accelerate the acquisition of crucial insights normally only available
  through \textit{ex-situ} part evaluation by means of machine learning analysis
  of \textit{in-situ} process monitoring data.
\end{abstract}

\section{Introduction}
Additive manufacturing (AM) presents a competitive alternative to the
conventional approaches in manufacturing with the advantages of efficient
material utilization, design consolidation, and fast iteration\cite{
rahman_review_2023, beaman_additive_2020, akbari_meltpoolnet_2022,
hemmasian_surrogate_2023}. However, a significant area of improvement lies
within defect prevention as printed parts present their own set of challenges in
porosity, distortion, and cracking\cite{rahman_review_2023}. These issues are
often uncovered through \textit{ex-situ} non-destructive testing methods and can
sometimes be addressed through lengthy post-processing means such as hot
isostatic pressing (HIPing) before they are certified
\cite{ordas_fabrication_2015, dolimont_effect_2016}. With \textit{in-situ}
process monitoring, a digital twin of the fabrication process can be created and
segments of the certification process can be conducted in
parallel\cite{knapp_building_2017, bartsch_digital_2021, gaikwad_toward_2020}.

Laser powder bed fusion (LPBF), relies primarily on established process maps
\cite{clymer_powervelocity_2017, agrawal_predictive_2022, grasso_-situ_2021} to
determine the optimal machine settings that minimize defects within the finished
part. Most commonly, these process maps explore the power and velocity space to
determine a combination of two that would result in a sufficiently dense part.
Informed control over these process parameters and others such as hatch spacing
\cite{xia_influence_2016}, layer height \cite{snyder_build_2015}, and raster
pattern \cite{mertens_optimization_2014}, can greatly affect the part's
porosity, microstructure \cite{gockel_understanding_2013}, and surface finish
\cite{snyder_build_2015}. However, even within build conditions with nominal
process parameters, defects such as porosity remain an issue
\cite{slotwinski_porosity_2014}.

\textit{In-situ} process monitoring offers a means to resolve this issue as
information obtained from the build process can assist in resolving many of the
technical challenges encountered during part fabrication
\cite{slotwinski_porosity_2014, noauthor_measurement_nodate, tian_roadmap_2022}.
Many of these defects and their precursors such as part distortions
\cite{biegler_-situ_2018}, surface roughness \cite{hofman_situ_2022}, or keyhole
formation \cite{ren_machine_2023, tempelman_detection_2022} exhibit signals
which with the appropriate sensors can be detected before \textit{ex-situ}
sample analysis. These indicators can be applied alongside the build process to
analytical and machine learning models in order to obtain the necessary feedback
to adjust process parameters for the build. This feedback loop would be
optimized to reduce the number of part defects through both preemptive and
responsive measures. \cite{mccann_-situ_2021, feng_predicting_2022}. In
addition, reconstructing the porosity map can significantly accelerate the part
certification process as the knowledge of the porosity map can expedite
qualification through observations of statistics alone\cite{seifi_overview_2016,
chen_review_2022, knapp_building_2017, dordlofva_design_2020,
sola_microstructural_2019}. 

Thermal imaging demonstrates effectiveness as an \textit{in-situ} process
monitoring technique as evidenced by previous studies which have explored
comparing melt pool images to computational fluid dynamics simulations
\cite{myers_high-resolution_2023}, mathematical equations
\cite{kayacan_investigation_2020}, and 3D surface maps\cite{haley_-situ_2021}.
Further exploration of this technique has shown effectiveness in applications
such as defect detection and correction within the build process either
indicating likely porosity given a thermal image of a melt pool
\cite{mitchell_linking_2020} or material extrusion correction in large scale
additive manufacturing\cite{borish_real-time_2020, borish_-situ_2019}.

Analytical solutions such as Rosenthal's equation
\cite{rosenthal_mathematical_1941} provide a foundational method to determine
nominal process parameter regions within laser power and scanning velocity
space. This equation can be adapted to provide depth and width estimates of the
melt pool given specific process parameters such as preheat temperature, power,
and velocity which can be applied to the selection of nominal parameters for
hatch spacing and layer height. However, this method poses limitations as
solutions provided by Rosenthal's equation are only suitable for melt pools
within the conduction regime \cite{rosenthal_mathematical_1941,
hekmatjou_comparative_2020, imani_shahabad_extended_2021}. This leaves areas out
that are not captured through analytical models such as melt pool behavior in
the keyhole mode and process conditions such as scan strategies.

Much attention has been directed towards machine learning to fill this gap
between the projection of these analytical models and their applied results some
of which include process parameter optimizing
\cite{baturynska_optimization_2018, ogoke_inexpensive_2023} and fatigue life prediction
\cite{zhan_machine_2021}. In this paper we explore the application of machine
learning towards the quantification and spatial localization of pores within a
sample given the \textit{in-situ} monitoring data of thermal images. These
predictions can then be utilized to create a digital twin of the built sample
and perform qualification and certification tasks in parallel to the sample
fabrication\cite{gaikwad_toward_2020,bartsch_digital_2021,knapp_building_2017}.

For the task of pore quantification a three dimensional Convolutional Neural
Network (CNN) was utilized to extract features within a provided sequence of
thermal images and provide a singular scalar prediction of the expected number
of pores.  Models such as \textit{ImageNet}\cite{krizhevsky_imagenet_2017} have
shown the effectiveness of 2D CNNs with image classification tasks and other
models such as \textit{C3D}\cite{tran_learning_2015} have applied this technique
to extended over a sequence of images. Training a 3D CNN model with the
objective of pore quantification enables the identification of pore counts
within a build layer prior to any \textit{ex-situ} sample analysis.

The task of pore localization utilizes a Video Vision Transformer
(ViViT)\cite{arnab_vivit_2021} which is suited to capture the spatial and
temporal features within the sequence of thermal images through subdividing the
input into patches. The original implementation of the ViViT model is structured
to provide a classification output\cite{arnab_vivit_2021}, however for the
purposes of pore localization the classification head is replaced with a dense
prediction head which retains the spatial information of the input sequence. Our
network implementation utilizes a dense output which directly correlates the
spatial and temporal information into a 2D pore localization prediction. This
network builds off the work by Ranftl et al.\cite{ranftl_vision_2021} where
fusion blocks and convolutional layers are added to a vision transformer to
provide depth predictions of a given image.

Additive manufacturing, specifically LPBF, relies on sample \textit{ex-situ}
post-build inspection and testing to qualify parts and identify potential
defects encountered during the build process\cite{damon_process_2018,
feng_additive_2022, chen_review_2022, donegan_multimodal_2021}. This often
includes tedious processes such as cross-sectional imaging or x-ray computed
micro-tomography (CT) after the part fabrication.  However, if similar
information is obtained earlier during the build process through the creation of
a digital twin, builds can be terminated earlier or dynamic adjustments can be
applied once the presence of defects is detected to reduce material waste and
costs.  Applying machine learning techniques such as CNN or ViViT models on
\textit{in-situ} data opens the possibility to acquire \textit{ex-situ} sample
insights during the build process and fabricate parts within a closed feedback
loop.

\begin{figure}[bt]
  \centering
  \includegraphics[width=\textwidth]{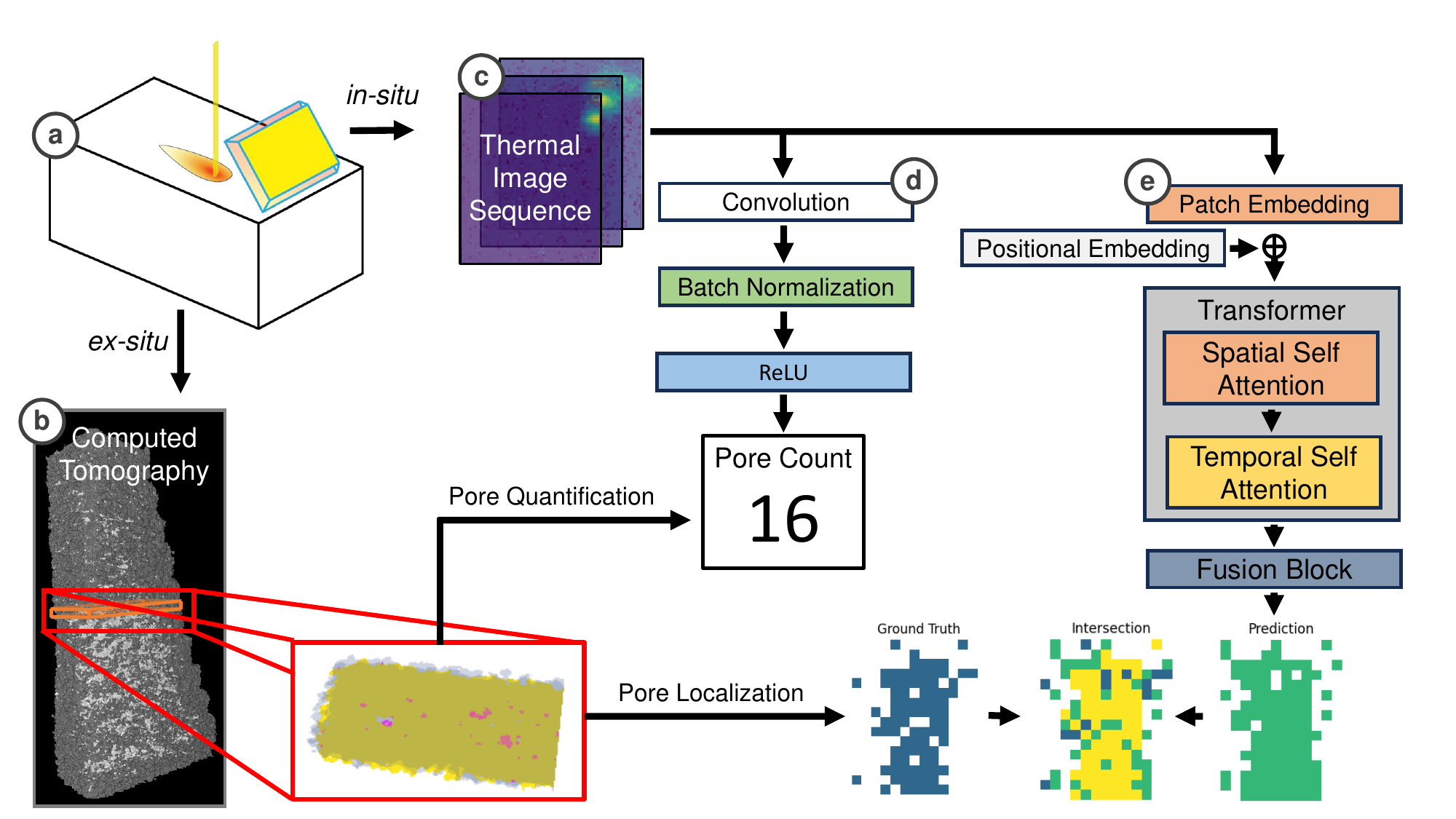}
  \caption{
    A sequence of 200 thermal pyrometry images (\ref{fig:main}c) providing
    absolute temperature values of the build plate taken \textit{in-situ}
    (\ref{fig:main}a) are provided as input data for models for pore
    quantification (\ref{fig:main}d) and pore localization (\ref{fig:main}e).
    These two separate models utilize a CNN and ViViT with dense output heads to
    produce a scalar number of pores and 2D mapping of expected porosity regions
    respectively. Metrics derived from \textit{ex-situ} CT data for the
    corresponding build layer are used as ground truth values for each model
    (\ref{fig:main}b).
  }
  \label{fig:main}
\end{figure}

\section{Methodology}

\subsection{\textit{Spacing} and \textit{Velocity} Samples}
\subsubsection{Sample Fabrication and Data Acquisition}
\begin{figure}[bt]
  \centering
  \begin{subfigure}{0.20\textheight}
      \centering
      \includegraphics[width=\textwidth]{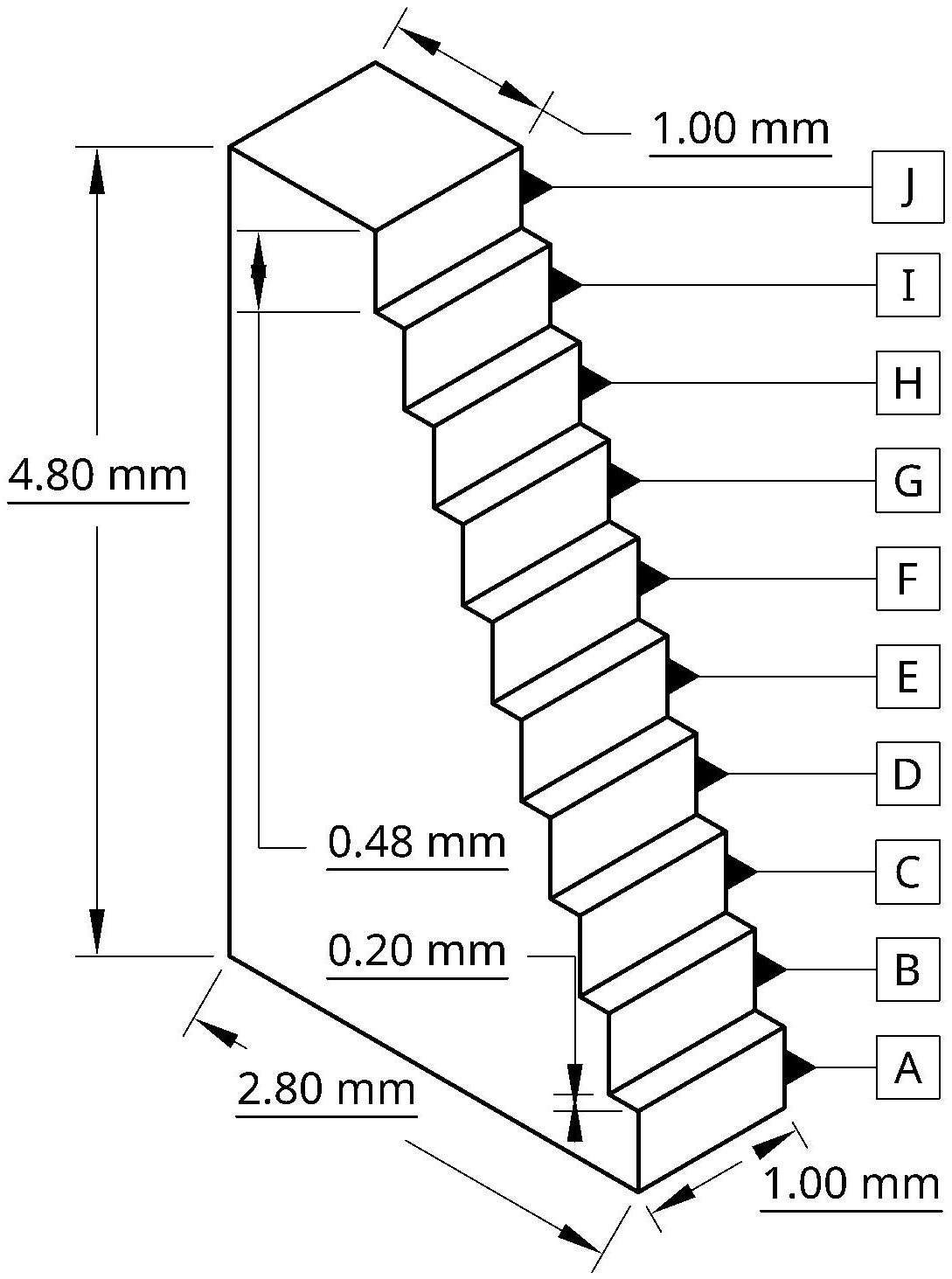}
      \caption{Sample Design}
      \label{img:sample_design}
  \end{subfigure}%
  \begin{subfigure}{0.33\textwidth}
      \centering
      \begin{tabular}{|c|c|}
          \hline
          \small\textbf{Step} & \small\textbf{Hatch Spacing} \\
          \hline
          J & 25 {\textmu}m \\
          \hline
          I & 30 {\textmu}m \\
          \hline
          H & 35 {\textmu}m \\
          \hline
          G & 40 {\textmu}m \\
          \hline
          F & 45 {\textmu}m \\
          \hline
          E & 75 {\textmu}m \\
          \hline
          D & 70 {\textmu}m \\
          \hline
          C & 65 {\textmu}m \\
          \hline
          B & 60 {\textmu}m \\
          \hline
          A & 55 {\textmu}m \\
          \hline
      \end{tabular}
      \caption{\textit{Spacing} Parameters}
      \label{tab:spacing_process_parameters}
  \end{subfigure}%
  \begin{subfigure}{0.33\textwidth}
      \centering
      \begin{tabular}{|c|c|}
          \hline
          \small\textbf{Step} & \small\textbf{Scan Velocity} \\
          \hline
          J & 1.05 m/s \\
          \hline
          I & 1.12 m/s \\
          \hline
          H & 1.19 m/s \\
          \hline
          G & 1.26 m/s \\
          \hline
          F & 1.33 m/s \\
          \hline
          E & 1.75 m/s \\
          \hline
          D & 1.68 m/s \\
          \hline
          C & 1.61 m/s \\
          \hline
          B & 1.54 m/s \\
          \hline
          A & 1.47 m/s \\
          \hline
      \end{tabular}
      \caption{\textit{Velocity} Parameters}
      \label{tab:velocity_process_parameters}
  \end{subfigure}
  \caption{
    Each sample (\ref{img:sample_design}) contains 10 different
    process parameter combinations  with the \textit{Spacing} sample
    (\ref{tab:spacing_process_parameters}) exhibiting varying hatch spacing
    and the \textit{Velocity} sample (\ref{tab:velocity_process_parameters})
    exhibiting varying scan velocity.
  }
  \label{fig:sample_fabrication}
\end{figure}

This paper analyzes two samples, one with variable hatch spacing
(\textit{Spacing}) and the other with variable scan velocity
(\textit{Velocity}). Both of these samples were manufactured using LPBF
equipment (ProX DMP 200 from 3D Systems) with AISI 316L stainless steel powder
and a constant laser power of 103 W\cite{arnhart_-situ_2022,
mitchell_linking_2020}.  These samples were designed with a staircase structure
(Fig.  \ref{img:sample_design}) with each sample comprised of 10 separate steps
and each step consisting of a 16 build layers with a 30 {\textmu}m layer height.
Within each of these steps a different combination of process parameters were
implemented with changes to either hatch spacing (Fig.
\ref{tab:spacing_process_parameters}) or scanning velocity (Fig.
\ref{tab:velocity_process_parameters}). The expected dimensions of each sample
are 4.80 mm \texttimes\;2.80 mm \texttimes\;1.00 mm in height, length, and width
respectively\cite{arnhart_-situ_2022}. Each step consisted of dimensions 0.48 mm
in height and ranged from 1.00 mm to 2.80 mm in length, decreasing in length by
0.20 mm from the top of the sample to the bottom. The \textit{Spacing} sample
with varying hatch spacing was built with a constant 1.4 m/s scan velocity and
the \textit{Velocity} sample with varying scan velocity was built with a
constant 50 {\textmu}m hatch spacing \cite{arnhart_-situ_2022}.A \textit{normal}
rastering pattern consisting of line scans parallel to the build axes,
orthogonal to the previous layer was utilized as the scan strategy for both
samples.

\subsubsection{In-situ Pyrometry}
Absolute temperature estimations
were calculated from thermal radiation captured by a Stratonics two-color
pyrometer receptive to light emitted at 750 nm and 900 nm, a frame rate within 6
- 7 kHz, and a 90 {\textmu}s exposure.\cite{mitchell_linking_2020}
Synchronization between the LPBF equipment and pyrometer were achieved via
Transistor-Transistor Logic (TTL) triggering producing 1000 frames of 65 px
{\texttimes} 80 px images within each build layer\cite{mitchell_linking_2020}.
This translated to a 1365 {\textmu}m {\texttimes} 1680 {\textmu}m resolution
with approximately 21 {\textmu}m per px. This presented a total of 159,000
images taken for each sample and with initial screening applied to filter out
"empty" images, reducing the total number of frames down to 20,469 and 20,187
for \textit{Spacing} and \textit{Velocity} samples respectively.

\subsubsection{Ex-situ Micro-computed Tomography (CT)}
Micro-computed tomography analysis was performed using a Zeiss Xradia 520 Versa
at the maximum output power of 10 W and tube voltage of 140 kV with the sample
positioned 11.1 mm from the source \cite{arnhart_-situ_2022}. Scans were taken
at a cubic voxel size of 3.63 \textmu m and with a build layer height of 30
\textmu m, this equated to approximately 8.26 voxels per build layer. The
obtained scans for the \textit{Spacing} and \textit{Velocity} samples were
bounded by 5.05 mm \texttimes\;3.00 mm \texttimes\;1.35 mm and 5.05 mm
\texttimes\;3.17 mm \texttimes\;1.44 mm respectively (Z \texttimes\;Y
\texttimes\;X) \cite{arnhart_-situ_2022}. Both \textit{Spacing} and
\textit{Velocity} samples extended 1410 voxels \texttimes\;900 voxels
\texttimes\;430 voxels along the Z, Y, and X axes. The scanned tomography was
further segmented to distinguish each pore from the general sample with
annotations denoting each of the labeled pore's centroid, equivalent diameter,
number of voxels, and unique identifier.

\begin{figure}[bt]
  \centering
  \includegraphics[width=\textwidth]{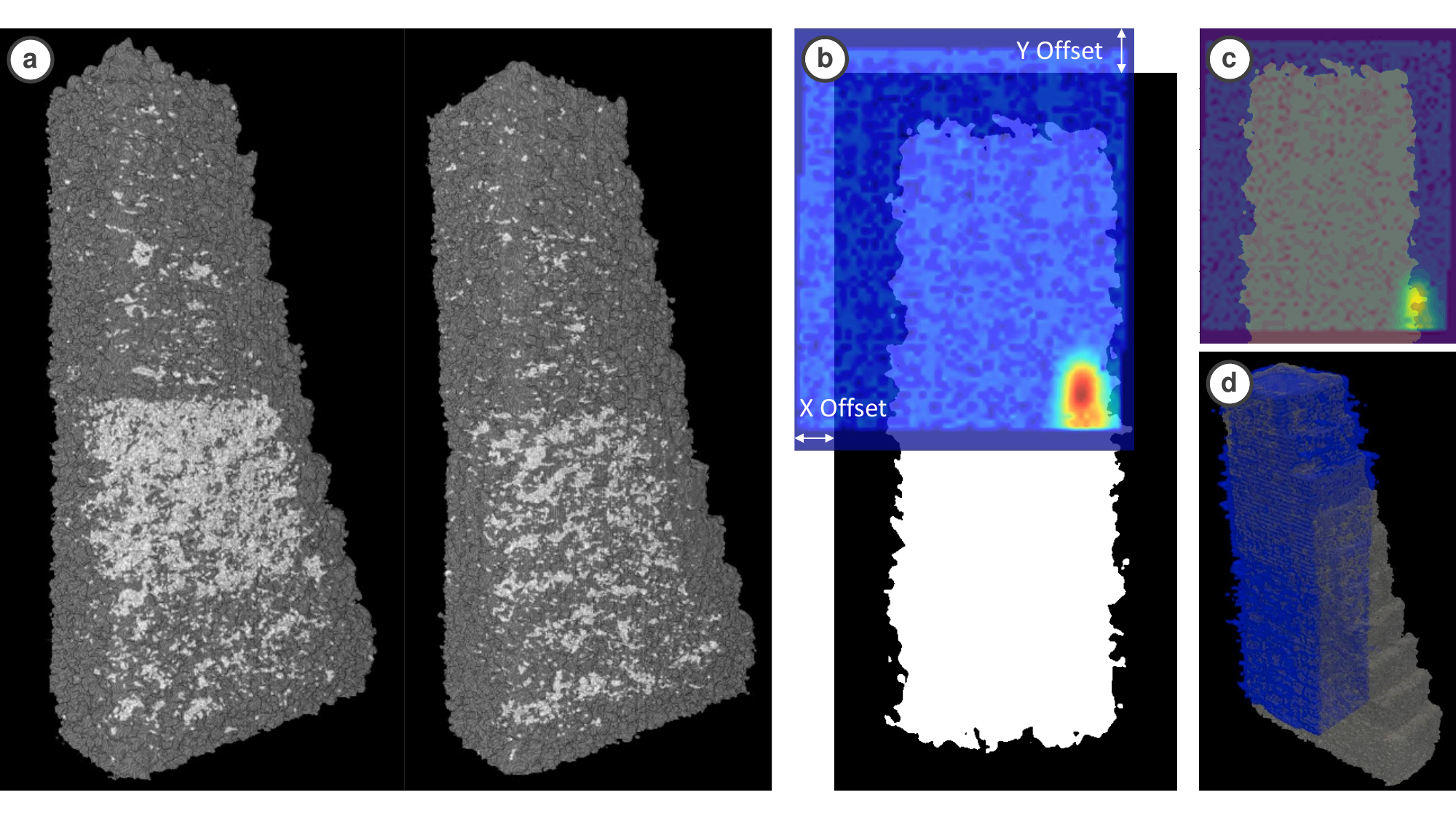}
  \caption{
    CT visualization of porosity segmented within overall sample for
    \textit{Spacing} (left) and \textit{Velocity} (right) samples
    (\ref{fig:sample_alignment}a). Pyrometry data is captured within a subset of
    the built sample (\ref{fig:sample_alignment}d) and initial alignment of
    image and CT data occurs along the X and Y axes
    (\ref{fig:sample_alignment}b), using the top left corner of both as the
    origin. Alignment for a build layer is visually validated with an overlay of
    the pyrometry data directly on top of a corresponding cropped CT voxel layer
    (\ref{fig:sample_alignment}c).
  }
  \label{fig:sample_alignment}
\end{figure}

\subsection{Dataset}
\subsubsection{Pyrometry and Micro-computed Tomography Alignment}
Capture area was limited to 80 px \texttimes\;65 px and lead to the raster
patterns of the lower build layers (steps A - F) to extend further than the
camera's viewport \cite{mitchell_linking_2020}. As mentioned, initial screening
filtered out many of these empty frames by applying a minimum threshold on each
frame's peak value from long wavelength data\cite{mitchell_linking_2020}. This
method removed the frames where the melt pool was out of view, however still
included some frames where spatter likely occurred.

\textit{In-situ} thermal images and slices of \textit{ex-situ} CT data shared
the same origin at the top left (Fig \ref{fig:sample_alignment} b). Small
offsets were then applied to align the X and Y directions of CT to the thermal
image. The provided Z direction offsets were used as starting points to align
the CT data to the corresponding build layer. The Z alignment for both samples
were visually verified through manual alignment of the scan path of thermal
images and the a top down view of the corresponding CT layer. Step G (Fig.
\ref{img:sample_design}) within both samples was the first section of the sample
where the entire scan path is in complete view of the thermal camera and was
used as a reference point to align the CT data. Both samples (\textit{Spacing}
and \textit{Velocity}) were offset by a total 5 build layers (\textasciitilde 9
voxels per build layer) in total.

\subsubsection{Pore Thresholding}
A CT resolution of 3.63 \textmu m per voxel allowed for the capture of distinct
shapes and contours associated with porosity, however this fine resolution also
resulted in recording scattered distributions of small voids. Accurately
predicting these small voids is a difficult task for a model as these defects
could be the result of gas porosity \cite{iebba_influence_2017} or rogue flaws
\cite{reutzel_application_2023} and may have precursors not visible to thermal
imaging. To achieve greater correlation between the pyrometry data and CT data,
our attention focused on larger diameter pores that can be attributed to factors
such keyhole porosity or lack of fusion porosity. In keyholing, pores generated
by the vapor column during builds resulted in an average diameter of 47 \textmu
m \cite{shrestha_study_2019} and lack of fusion pores with diameters dependent
on the height and width of melt pool and corresponding build layer
\cite{cacace_lack_2022}.

\begin{figure}[bt]
  \centering
  \includegraphics[width=\textwidth]{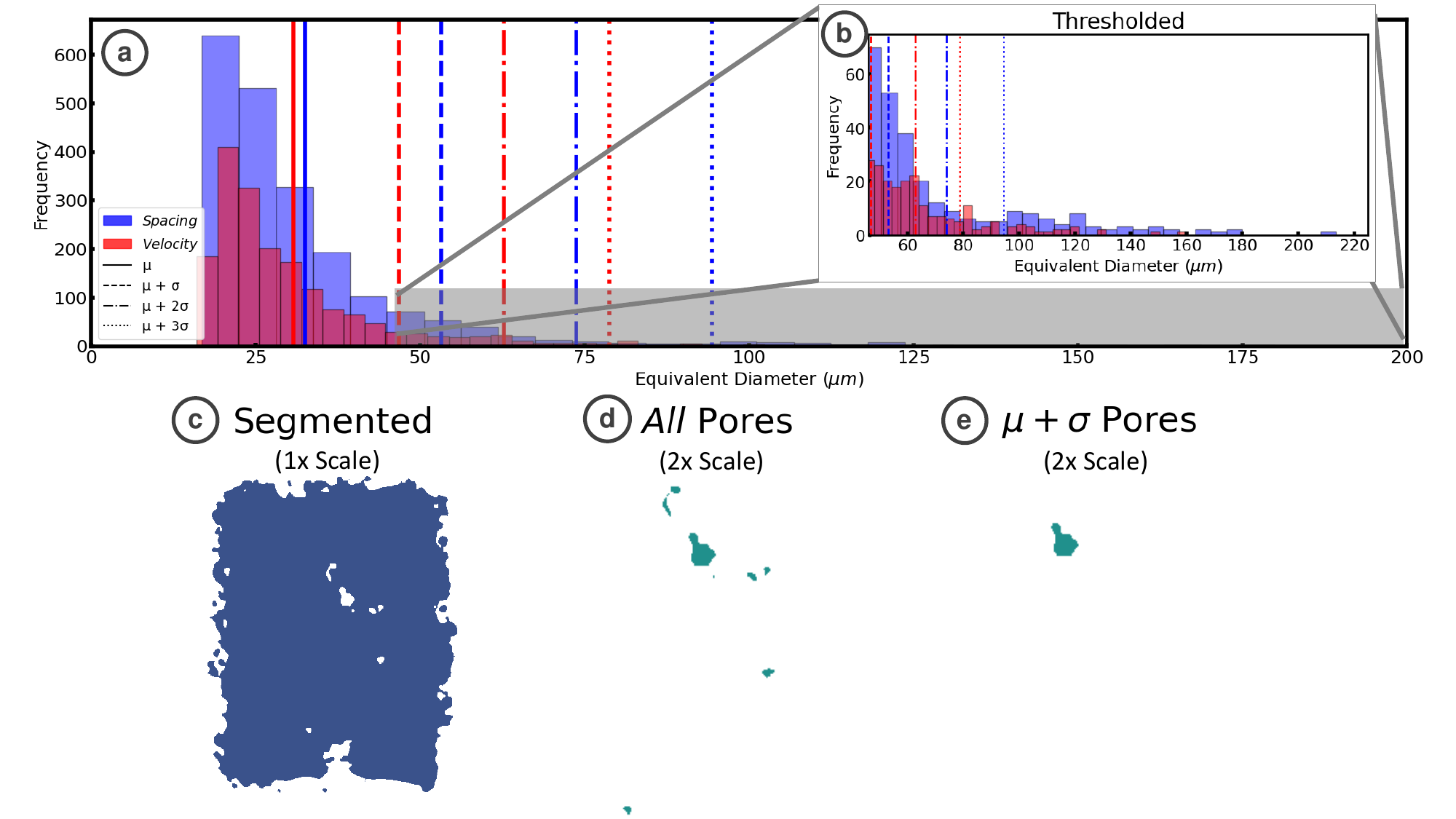}
  \caption{
    The \textit{Spacing} sample on average consist of larger pores with a
    standard deviation of 20.5 \textmu m compared to the \textit{Velocity}
    sample's standard deviation of 16.0 \textmu m
    (\ref{fig:porosity_threshold}a). The tighter distribution of
    \textit{Velocity} pores are visible in the thresholded distribution of
    equivalent pore sizes (\ref{fig:porosity_threshold}b) within the slices
    of the CT data. The segmentation of pores within the sample
    (\ref{fig:porosity_threshold}c) relied on a minimum of 100 voxels (11.4
    \textmu m equivalent spherical diameter)\cite{mitchell_linking_2020} in
    order to register the contiguous cluster of voids as porosity
    (\ref{fig:porosity_threshold}d) and an increase in the minimum size
    further removed smaller pores (\ref{fig:porosity_threshold}e).
  }
  \label{fig:porosity_threshold}
\end{figure}

The mean Equivalent Spherical Diameter (ESD) for each sample was compiled in
order to obtain minimum threshold values 1 standard deviation above the mean. In
the \textit{Spacing} sample, pores exhibited an average diameter of 32.59
\textmu m and a standard deviation of 20.60 \textmu m, resulting in a minimum
(\textit{$\mu + \sigma$}) ESD threshold of 53.19 \textmu m.  In the
\textit{Velocity} sample, pores had an average diameter of 30.78 \textmu m and a
smaller standard deviation of 16.01 \textmu m resulting in a minimum
(\textit{$\mu + \sigma$}) ESD thresholds of 46.79 (Fig.
\ref{fig:porosity_threshold}b).

\section{Porosity Reconstruction}
The quantity of pores and their position within the build layer was
reconstructed with machine learning inferencing upon a sequence of thermal
images.  With the set of \textit{in-situ} and \textit{ex-situ} data, the tasks
of predicting the number of pores and the approximate location of these pores
within the various build layers were achieved using a CNN model and a ViViT
model with dense prediction heads respectively.

The number of pores corresponding to a sequence of thermal images were predicted
using a CNN model. In this task both the application of rotational transforms on
the input sequence and the volumetric depth utilized for pore count were treated
as variables. Labels were derived from counting the set of unique pores within a
specified volumetric depth corresponding to 1, 2 or 3 build layers and
duplication was avoided by cross referencing each voxel's pore id. The quality
of these predictions is measured using Root Mean Square Error (RMSE) and
$\text{R}^2$ score.

The localization of pores was predicted using the same sequence of thermal
images and utilized a ViViT model with a dense prediction head to indicate
sections expected to be porous. The labels for this task were obtained by
downsampling the CT data for the build layer equally along all axes in order to
provide a coarse porosity estimates for the model to train and predict. In
addition, the effect of applying rotational transforms on the input sequence and
minimum pore ESD thresholds for label compilation were investigated. The quality
of the predictions was measured using an Intersection over Union (IoU) score,
considering the overlap of the area of predicted porosity over that of the
label.

\subsection{Porosity Count}
This task investigates the extent in which sequences of thermal images can
quantify the number of pores that exist within the sample build layers. The
\textit{Spacing} sample consists of 2069 pores and the \textit{Velocity} sample
consists of 1811 pores. The labels were obtained from slicing the CT sample data
into the appropriate dimensions that corresponding to the build layer and
counting the number of unique pore identifiers omitting the 0 value which
represented the background. Each build layer consisted of 9 voxels in depth and
this volume was extended to a depth of 18 and 27 voxels to obtain pore counts
extending down 2 and 3 build layers.

\begin{figure}[bt]
  \centering
  \includegraphics[width=\textwidth]{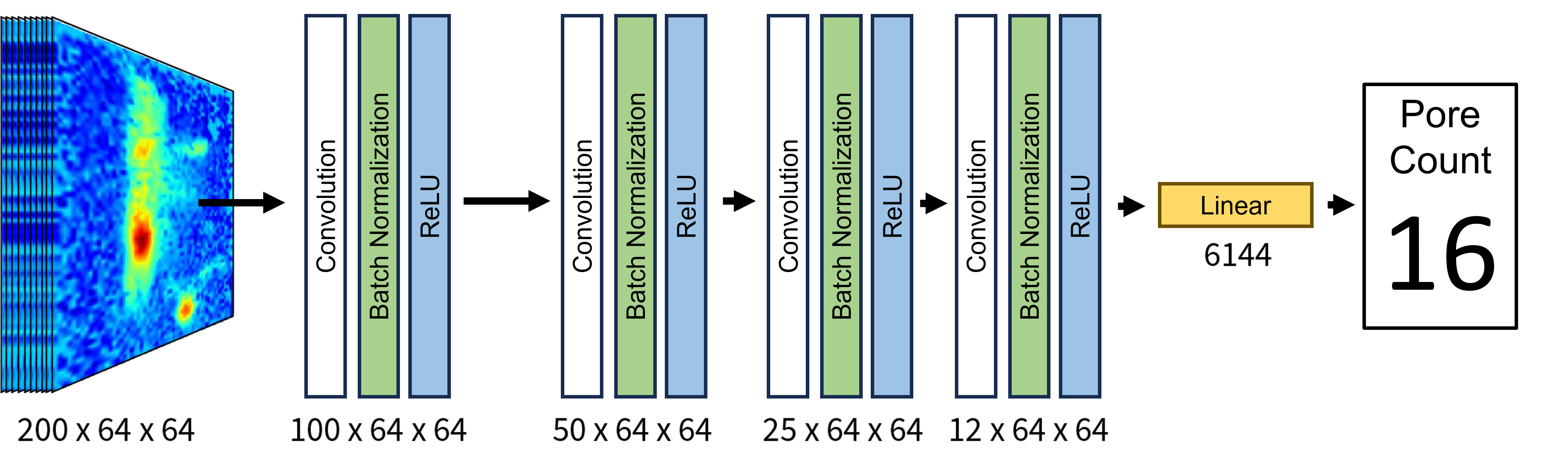}
  \caption{
    A standard convolutional neural network with a kernel size of 3 pixels and a
    stride of 2 filters the features within an image down over 4 layers to a
    singular scalar value indicating the expected number of pores from the
    sequence of thermal images.
  }
  \label{fig:cnn}
\end{figure}

In this task a CNN model composed of 4 convolutional layers and 2 fully
connected layers reduce the input set of 200 64 \texttimes\;64 pixel images into
a scalar value of the number of unique pores within the build layer (Fig.
\ref{fig:cnn}). A 3 \texttimes\;3 pixel kernel is convolved on top each image
with a stride of 2 and a padding of 1. Within each layer the number of channels
is reduced by a factor of 2 and batch normalization and ReLU non-linearity
activation function are applied. The output of the CNN layers are reshaped into
a 2 dimensional tensor before they are passed to the two fully connected layers
which output a single scalar value which quantifies the number of pores within
the build layer.

A CNN operates by leveraging convolutional layers to extract features from input
images hierarchically. In the initial layers, low level features such as edges
and gradients are detected through convolutional operations where the kernel
moves across the input image, computing dot products and producing feature maps
\cite{krizhevsky_imagenet_2017, tran_learning_2015, long_fully_2015,
jadhav_stressd_2023}. Activation functions such as ReLU apply non-linear
transforms and allow for the capture of complex patterns. The following layers
then build upon these low level feature maps and repeat this task of
transforming the raw pixel values into the outputs for a specific task
\cite{krizhevsky_imagenet_2017}. These tasks are determined by the final layers
of the network which utilize fully connected layers to shape the output to
perform classification, regression, or reconstruction.

Due to dataset size constraints, a 75 / 25 train test split of the data was used
for model training. The train test split occurred within the 16 build layers of
each of the 10 sample steps of either the \textit{Spacing} or \textit{Velocity}
sample. This split within sample steps was implemented to provide an equal
distribution of processing parameters between the train and test sets for the
model. This provided an input label set of 120 training pairs and 39 testing
pairs for the either of the samples. Our dataset implementation allowed for the
model to train on either the \textit{Spacing}, \textit{Velocity}, or on a
combination of both datasets with the \textit{All} dataset.  For this regression
task, a mean squared error was utilized as the loss function and the predicted
value from the model is rounded to the nearest integer. Each of the models were
trained from 500 epochs with a learning rate of 0.0001 using the ADAM optimizer.

In addition, data augmentations of the input sequence in the form of rotational
transforms for the entire video sequence were applied. Data augmentation
provides a means to artificially expand an existing set of data in order to
improve the generalization ability and robustness of the
model\cite{krizhevsky_imagenet_2017}. Typical transformations change the input
image applying one or a combination of rotations, translations, flips, scaling,
and cropping. In this application we focus only on applying rotational
transforms ranging between $0^{\circ}$ to $180^{\circ}$ and avoid alterations to
the contrast or brightness that would affect the raw temperature value.

\subsection{Porosity Localization}
The localization task identifies areas within the build layer where pores are
likely to form through analyzing a sequence of thermal images. A video vision
transformer (ViViT) provides an applicable architecture to thoroughly analyze
the series of input frames to extract positional features through use of spatial
and temporal attention.\cite{arnab_vivit_2021} In our model implementation a
sequence of thermal images is provided to the model to map to localized porosity
labels obtained through alignment and extraction of the CT data. These build
layers of initial size of 9 \texttimes\;423 \texttimes\;520 voxels are cropped
and downsampled by a factor of 24 to a coarser 2 dimensional 1 \texttimes\;16
\texttimes\;16 voxel shape to provide a general area in which porosity is
expected.

\begin{figure}[bt]
  \centering
  \includegraphics[width=\textwidth]{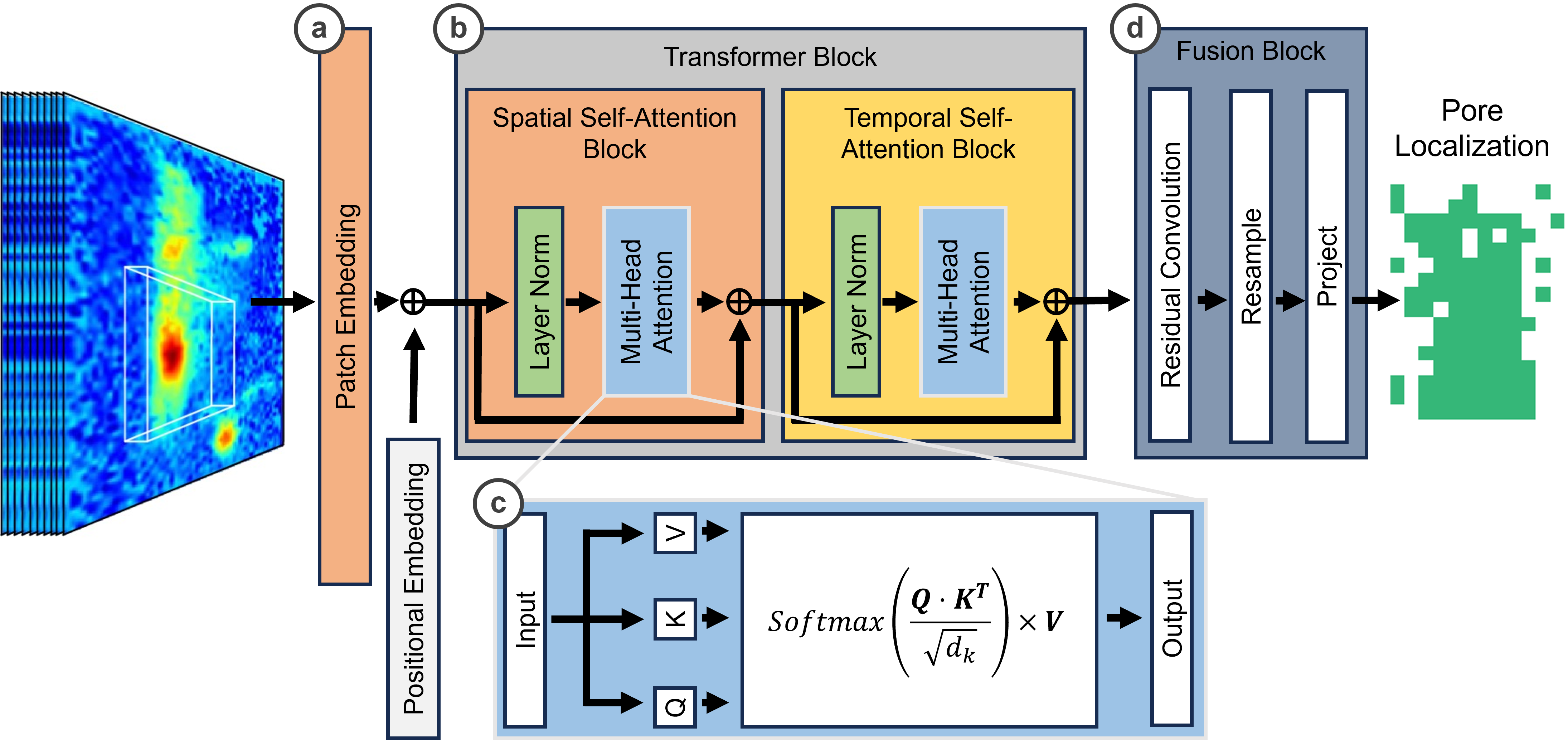}
  \caption{
    The input sequence of thermal images are sliced into a set of patches
    (\ref{fig:vivit}a) where 4 and 5 sublayers of the respective spatial and
    temporal self attention are applied (\ref{fig:vivit}b). Each self attention
    block consisted of 8 attention heads (\ref{fig:vivit}c) of a dimension of
    256. A feature fusion block (\ref{fig:vivit}d) applies residual convolution
    and a dense prediction head produces a 2 dimensional output indicating
    regions of expected porosity.
  }
  \label{fig:vivit}
\end{figure}

For this task a video vision transformer model with a dense prediction output is
utilized to localize pores within the sample space. This model is composed of a
spatial transformer layer with 4 sub-layers and a temporal transformer layer
with 5 sub-layers both with 8 heads and a dimension of 256 (Fig.
\ref{fig:vivit}). Afterwards the class tokens resulting from each of the
transformer layers are removed and the output is passed into a feature fusion
block which performs residual convolution and provide a fine grain prediction. A
series of 4 convolutional layers and ReLU non-linearity layers are applied
before passing to a sigmoid activation function to indicate whether porosity is
expected at a voxel location.

Attention within a transformer is comprised of three learnable components: The
query vector $\{\textbf{q}_i\}^{N_q}_{i=1}$, key vector
$\{\textbf{k}_i\}^{N_k}_{i=1}$, and the value vector
$\{\textbf{v}_i\}^{N_v}_{i=1}$ given that $N_k = N_v$\cite{bahdanau_neural_2016,
vaswani_attention_2023, li_scalable_2024}. In the attention mechanism, the query
vector retrieves contextual information from the key vector and generates an
output based on the weighted sum of corresponding value vectors. Contextual
information is retrieved in the form of an attention score which is the scaled
dot product between the query vector and the key vector:
$\frac{\textbf{q}_i\cdot\textbf{k}_j^T}{\sqrt{d}}$\cite{vaswani_attention_2023}.
The attention score is then used in the calculation of weights ($\alpha_{ij}$)
which apply a Softmax over the individual contribution of each attention score
(Eq. \ref{eq:attention}) \cite{vaswani_attention_2023, bahdanau_neural_2016,
li_scalable_2024}.Each token's numerical encoding along with its relevance to
other tokens is calculated from the cross product between the value vector and
weights resulting in the attention mechanism:
$Softmax\left(\frac
{\textbf{Q} \cdot \textbf{K}^\textbf{T}}
{\sqrt{d_k}}
\right)\times V$
\cite{vaswani_attention_2023, bahdanau_neural_2016, li_scalable_2024}.

\begin{equation}
\alpha_{ij} = Softmax\left(\frac
{e^{\frac{\textbf{q}_i\cdot\textbf{k}_j^T}{\sqrt{d}}}}
{\sum^N_{k=1}e^{\frac{\textbf{q}_i\cdot\textbf{k}_j^T}{\sqrt{d}}}}
\right) = Softmax\left(\frac
{\textbf{Q} \cdot \textbf{K}^\textbf{T}}
{\sqrt{d_k}}
\right)
\label{eq:attention}
\end{equation}

In the case of a Vision Transformer (ViT), an image is divided up into self
attention patches via patch embedding, passed through the transformer encoder,
and utilized through a classification head\cite{dosovitskiy_image_2021,
arnab_vivit_2021}.  Within patch embedding, each patch has a fixed pixel size in
height and width and its embedding is derived through flattening and linear
projection\cite{dosovitskiy_image_2021, arnab_vivit_2021}.  Class and position
are applied to the embedding before passed through the transformer encoder
composed blocks of layer norm, multi-head attention, and MLP layers after which
a classification head is attached.\cite{dosovitskiy_image_2021}

The dataset utilizes the same 75 / 25 train test split within the sample steps
for all the \textit{Spacing}, \textit{Velocity}, and \textit{All} variants. The
model was trained for 1000 epochs utilizing a binary cross entropy loss function
for the binary prediction of a voxel's porosity classification. An ADAM
optimizer along with a cosine decay learning rate scheduler with an initial 10
epoch warm up period from learning rates 0.00001 to 0.0001 is applied to help
with regularization and stabilization \cite{gotmare_closer_2018}. 

\section{Results and Discussion}

\subsection{Porosity Count}
The number of pores within a volume of the sample was predicted from a sequence
of thermal images using a CNN model trained on either the \textit{Spacing},
\textit{Velocity}, or \textit{All} dataset. Hyperparameters such as the
utilization of rotational transforms and the number of build layers involved in
the compilation of the pore count were investigated.

\begin{figure}[bt]
  \centering
  \begin{subfigure}[b]{0.3\textwidth}
    \centering
    \includegraphics[width=\textwidth]{
      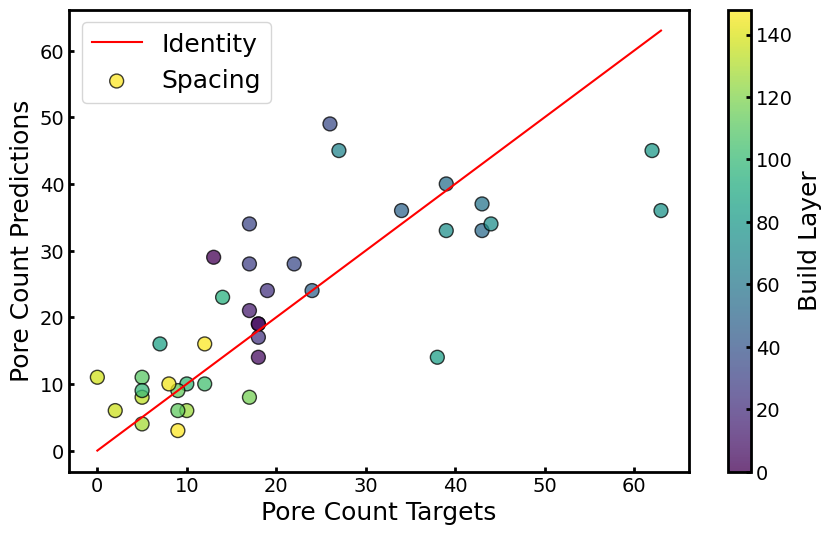
    }
    \caption{\textit{Spacing}}
    \label{fig:spacing_not_rotated}
  \end{subfigure}
  \hfill
  \begin{subfigure}[b]{0.3\textwidth}
    \centering
    \includegraphics[width=\textwidth]{
      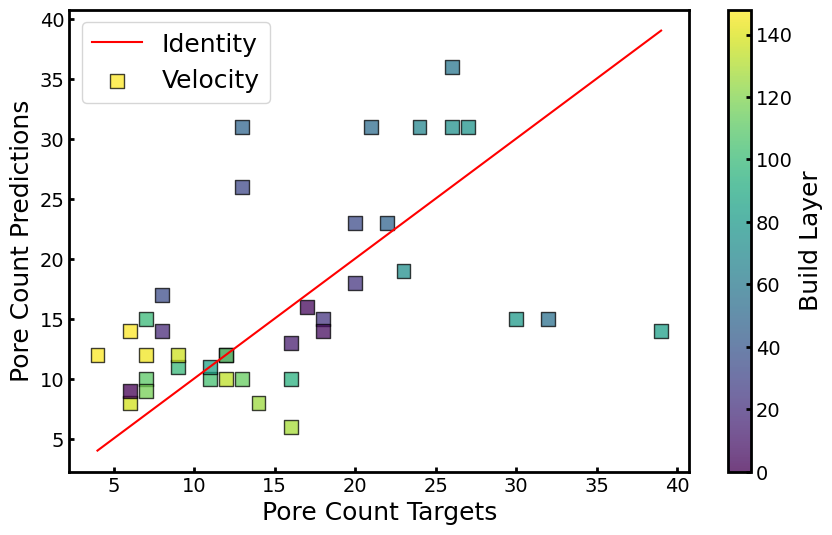
    }
    \caption{\textit{Velocity}}
    \label{fig:velocity_not_rotated}
  \end{subfigure}
  \hfill
  \begin{subfigure}[b]{0.3\textwidth}
    \centering
    \includegraphics[width=\textwidth]{
      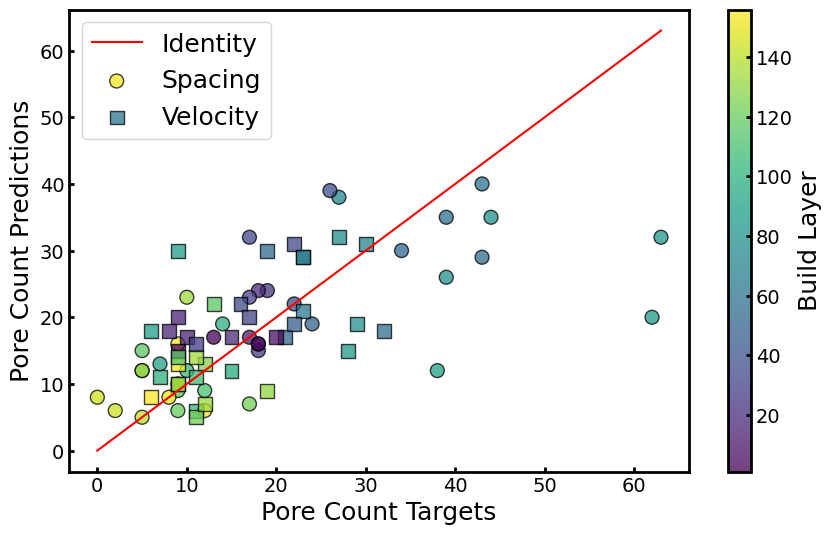
    }
    \caption{\textit{All}}
    \label{fig:all_not_rotated}
  \end{subfigure}
  \\
  \begin{subfigure}[b]{0.3\textwidth}
    \centering
    \includegraphics[width=\textwidth]{
      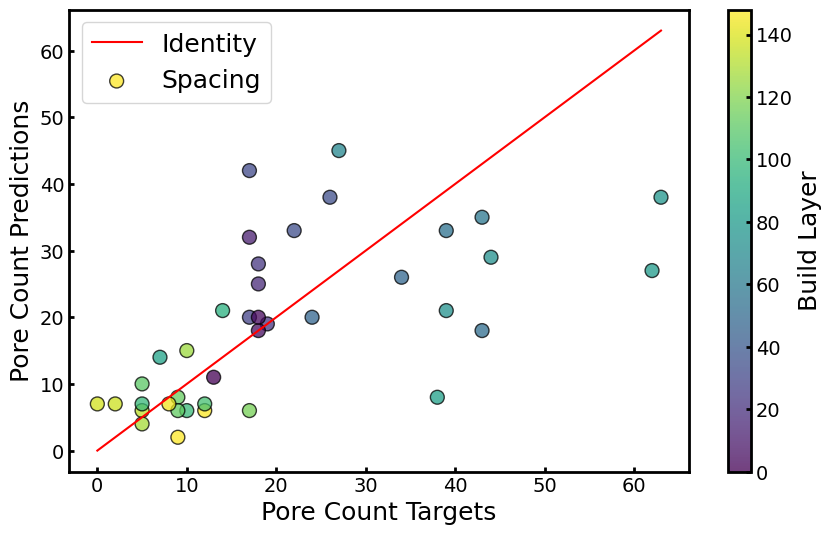
    }
    \caption{\textit{Spacing} (with transforms)}
    \label{fig:spacing_rotated}
  \end{subfigure}
  \hfill
  \begin{subfigure}[b]{0.3\textwidth}
    \centering
    \includegraphics[width=\textwidth]{
      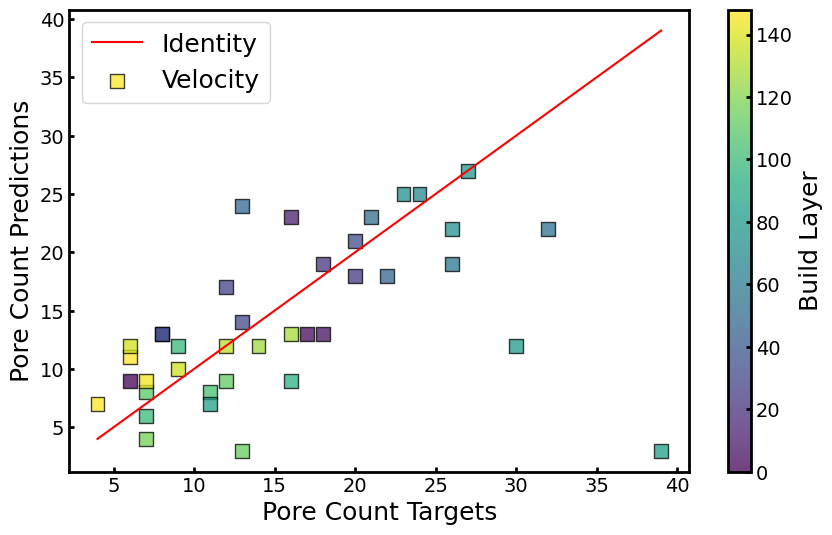
    }
    \caption{\textit{Velocity} (with transforms)}
    \label{fig:velocity_rotated}
  \end{subfigure}
  \hfill
  \begin{subfigure}[b]{0.3\textwidth}
    \centering
    \includegraphics[width=\textwidth]{
      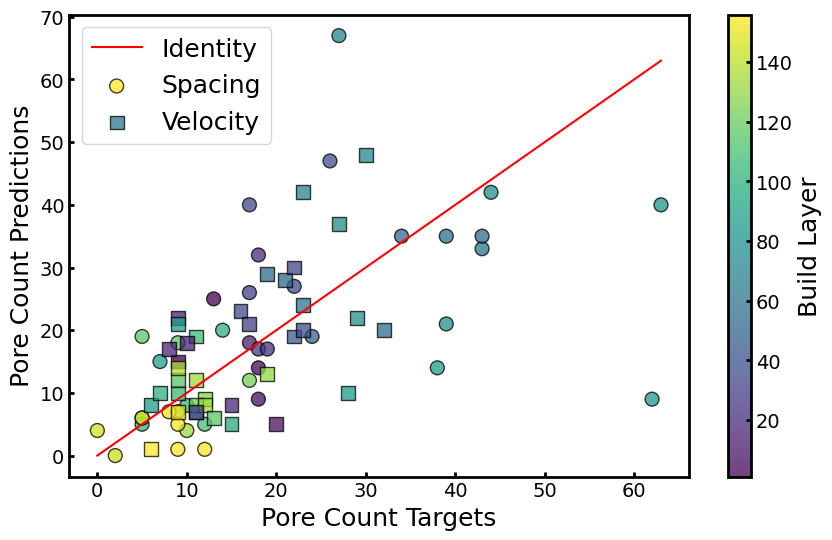
    }
    \caption{\textit{All} (with transforms)}
    \label{fig:all_rotated}
  \end{subfigure}
  \caption{
    CNN pore count predictions plotted alongside target pore values for
    \textit{Spacing}, \textit{Velocity}, and \textit{All} (combination of both
    datasets) for a depth of 1 build layer. In all plots the pore counts for
    build layers 100 - 160 are clustered near the origin as the upper half of
    the sample is composed of nominal process conditions. Of all the various
    implementations, the CNN model trained on the \textit{Spacing} (Fig.
    \ref{fig:spacing_not_rotated}) dataset without transforms shows the greatest
    level of alignment between the prediction and ground truth.
  }
  \label{fig:pore_count_results}
\end{figure}

Table \ref{tab:pore_count_cnn_results} outlines the model's performance
according to its training dataset and indicates the data augmentation procedures
that were applied and the various depths used to calculate pore count. Models
with datasets spanning 1 build layer exhibited the lowest error and highest
$\text{R}^2$ with a minimum RMSE score of 7.84 from the \textit{Velocity} sample
and a maximum $\text{R}^2$ score of 0.57 from the \textit{Spacing} sample. The
RMSE and $\text{R}^2$ score aim to measure the average magnitude of errors and
capture the proportion of variance within the model respectively. The alignment
of the prediction and the corresponding target for each model (Fig.
\ref{fig:pore_count_results}) shows a general trend where there is a larger
degree of porosity within the build layers ranging from 60 to 100 and to a
lesser degree elsewhere. Notably, build layers greater than 100 display lower
amounts of porosity for all models regardless of dataset or data augmentation.

This is expected as both samples transition from off nominal process parameters
to nominal process parameters towards the middle of the sample (Table
\ref{tab:spacing_process_parameters}, \ref*{tab:velocity_process_parameters})
with the upper portion of each sample fabricated with ideal process parameters.
For all models, areas where the sample was fabricated with nominal process
parameters show a greater degree of clustering as pores are sparse and few. In
earlier build layers, specifically those closer towards the middle of the sample
a greater spread of predictions is seen (Fig. \ref{fig:velocity_not_rotated},
\ref{fig:velocity_rotated}, \ref{fig:all_rotated}).

The model trained with the \textit{Spacing} dataset without rotational
transforms (Fig. \ref{fig:spacing_not_rotated}) achieved the highest
$\text{R}^2$ score of all models with a score of 0.57. This indicates that the
more than half of the variance within this model is able to be explained by the
input data. However, the model trained with the \textit{Velocity} sample and
rotational transforms achieved the lowest RMSE score of 7.84 of all models on
input data with significantly less variability as indicated with their lower
$\text{R}^2$ scores. The model trained on the \textit{All} dataset produced RMSE
and $\text{R}^2$ scores inbetween that of models trained on either
\textit{Spacing} or \textit{Velocity} datasets except in the situation for the
case where rotational transforms were applied where it yielded a $R^2$ score of
0.07. The additional hyperparameter of various build layer depths displayed
worse RMSE and $\text{R}^2$ scores indicating that there exists a greater
correlation between the thermal images and the build layer directly underneath
it.

The RMSE and $\text{R}^2$ values align with what is expected of the two
\textit{Spacing} and \textit{Velocity} datasets as the hatch spacing and scan
velocity are the two variables that change between sample steps respectively. In
the case of \textit{Spacing} sample, the hatch spacing produces a visible signal
in the form total rasters that is visible over the sequence of input frames.
However, the \textit{Velocity} sample uses a consistent number of rasters
traveling both vertically and horizontally across the build plate. The most
significant visual signal in the \textit{Velocity} sample is the distance the
melt pool travels inbetween frames.

\begin{table}
  \begin{tabular}{ll|ll|ll|ll}
    \hline
    \rowcolor{gray!30}
                        &                   & \multicolumn{2}{|l}{\textit{1 Build Layer}}   & \multicolumn{2}{|l}{\textit{2 Build Layers}}  & \multicolumn{2}{|l}{\textit{3 Build Layers}}  \\
    \rowcolor{gray!30}
    Dataset             & Data Augmentation & RMSE            & $\text{R}^2$                & RMSE            & $\text{R}^2$                & RMSE            & $\text{R}^2$                \\
    \hline                                                                                                                      
    \textit{Spacing}    &  Rotational       & 12.60           & 0.33                        & 19.04           & 0.18                        & 22.94           & 0.17                        \\
                        &  None             & 10.14           & \textbf{0.57}               & 18.46           & 0.23                        & 19.75           & \textbf{0.39}               \\
    \rowcolor{gray!10}                                                                                                          
    \textit{Velocity}   &  Rotational       & \textbf{7.84}   & 0.09                        & 11.41           & -0.04                       & 14.74           & 0.03                        \\
    \rowcolor{gray!10}                                                                                                          
                        &  None             & 8.08            & 0.03                        & \textbf{8.95}   & \textbf{0.36}               & \textbf{13.25}  & 0.22                        \\
    \textit{All}        &  Rotational       & 11.90           & 0.07                        & 13.89           & 0.34                        & 17.81           & 0.26                        \\
                        &  None             & 9.73            & 0.38                        & 14.57           & 0.27                        & 16.88           & 0.34                        \\
    \hline
  \end{tabular}
  \caption{
    $\text{R}^2$ and RMSE prediction performance metrics for CNN model trained
    on various datasets and build layer depths for 500 epochs.
  }
  \label{tab:pore_count_cnn_results}
\end{table}

\subsection{Porosity Localization}

During the training process, rotation transforms of the entire video sequence
were introduced as data augmentation methods and compared against models trained
without any rotational transformations. In addition to data augmentations, the
datasets were adjusted to allow for training on \textit{All Pores} and on pores
with ESD 1 standard deviation above the mean (\textit{$\mu + \sigma$ Pores}) in
an effort to investigate the model's performance on identifying the larger pores
within the dataset.

The position of the pores within the build layer was predicting using a sequence
of thermal images with the ViViT Dense model trained on either the
\textit{Spacing}, \textit{Velocity}, and \textit{All} dataset. The effect of
data augmentation and applying a threshold for minimum ESD for
pores were investigated with the training of this model as well. The
Intersection over Union (IoU) also known as the Jaccard Index was used to
quantify the performance of each model's prediction. The intersection over union
quantifies the prediction area overlap onto the target with the highest metric
of 1.0 occurring from an exact overlap of the two sets. For the IoU calculation
each set only includes the areas of porosity and in the case where both the
target and prediction exhibited no porosity, a score of 1.0 was given as the
prediction provided an exact match of the ground truth. 

\begin{figure}[bt]
  \centering
  \begin{subfigure}[b]{0.45\textwidth}
    \centering
    \includegraphics[height=\textheight/3]{
      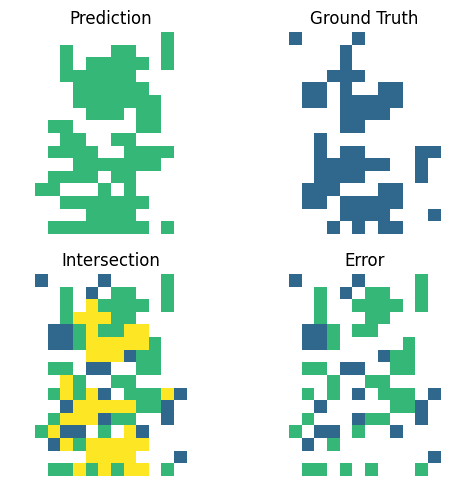
    }
    \caption{\textit{IoU Overlay with \textit{All Pores}}}
    \label{fig:iou}
  \end{subfigure}
  \begin{subfigure}[b]{0.45\textwidth}
    \centering
    \includegraphics[height=\textheight/3]{
      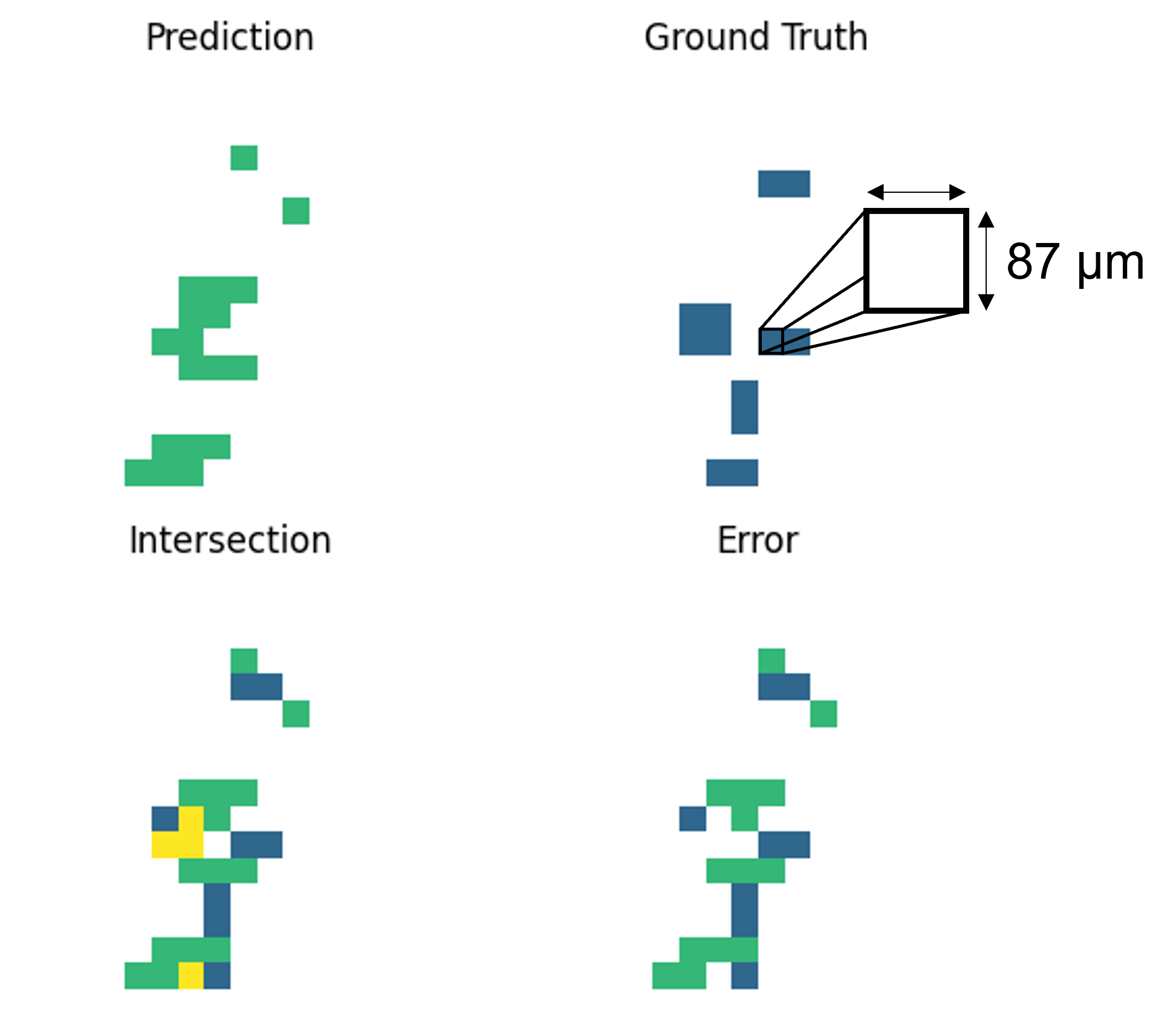
    }
    \caption{\textit{IoU Overlay with \textit{$\mu + \sigma$ Pores}}}
    \label{fig:iou_threshold}
  \end{subfigure}
  \caption{
    Comparison of prediction and label values for the model trained on the
    \textit{All} sample alongside the application of a minimum ESD pore
    threshold (Fig. \ref{fig:iou_threshold}). The areas colored in yellow
    represent the intersection that is considered for IoU calculations.
  }
  \label{fig:ious}
\end{figure}

The effect of training on the various \textit{Spacing}, \textit{Velocity}, and
\textit{All} datasets were investigated for this task as well the impact of 
rotational transforms on the input sequence. In addition, the prediction
performance of the model upon applying a threshold to hide pores smaller than 1
standard deviation above the mean was applied. The prediction results of these
models measured for each input label pair within the test dataset with the
overall average and maximum IoU scores recorded on Table
\ref{tab:pore_localization_vivit_dense_results}. Within the datasets that
included \textit{All Pores}, models trained on the \textit{Spacing} datasets
performed the best on average IoU scores when trained without rotational
transforms. A maximum IoU score of 0.85 was achieved within the model trained on
the dataset for \textit{All} samples with rotational transforms applied. Without
the applied threshold for minimum pore ESD, the model trained on the
\textit{Velocity} produces the lowest IoU score of the three models. However,
after a minimum pore ESD threshold is applied, the \textit{Velocity} dataset
model performs on par to that of the model trained with the \textit{Spacing}
dataset.  In all of the cases there was a greater trend in overlap within the
lower layers of the sample (Fig. \ref{fig:iou}) likely due to the higher
porosity resulting from the off nominal process conditions used within those
sample steps.

With a minimum pore threshold of 1 standard deviation above the mean pore
equivalent diameter (\textit{$\mu + \sigma$ Pores}), all models were able to
achieve maximum IoU scores of 1. These results were achieved towards the top of
each sample where nominal process conditions were used and correctly predicted
the ground truth of no pores above the threshold were present. (Fig.
\ref{fig:iou_threshold}) In some cases, areas with higher levels of porosity
produced lower IoU scores after thresholding as the previously larger regions
associated with porosity are reduced in size (Fig. \ref{fig:ious}). Overall, the
mean IoU increased for all models after the application of a minimum pore ESD
threshold.  The greatest of these increases is seen in the models trained with
the \textit{All} dataset of which scored the highest average IoU of any other
models. Rotational transforms did not prove to have a significant impact on
improving training as the resulting metrics were often within a percent error
from each other.

\begin{figure}[bt]
  \centering
  \begin{subfigure}[b]{0.3\textwidth}
    \centering
    \includegraphics[width=\textwidth]{
      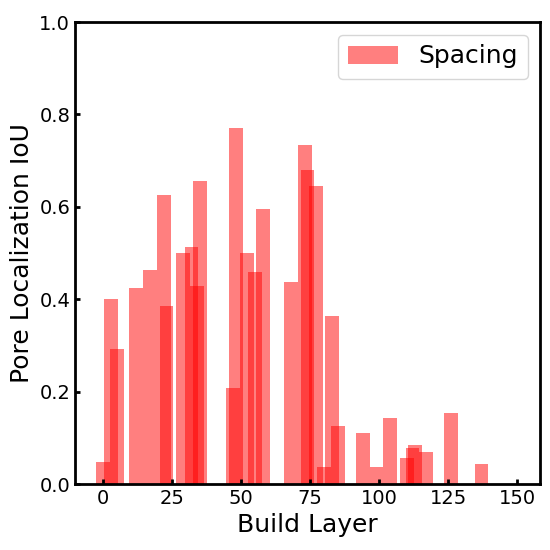
    }
    \caption{\textit{Spacing}}
    \label{fig:localization_spacing_not_rotated}
  \end{subfigure}
  \hfill
  \begin{subfigure}[b]{0.3\textwidth}
    \centering
    \includegraphics[width=\textwidth]{
      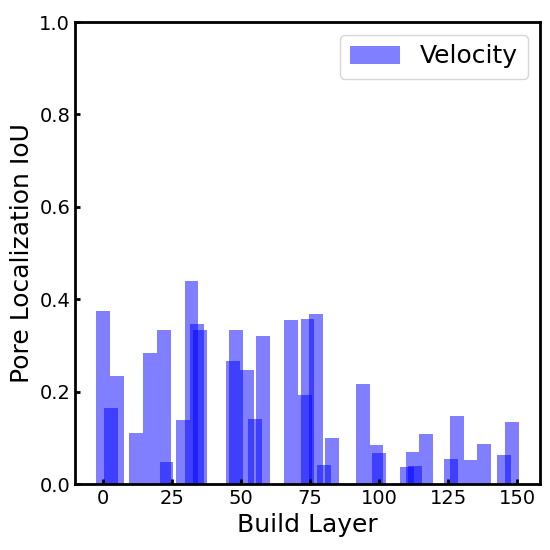
    }
    \caption{\textit{Velocity}}
    \label{fig:localization_velocity_not_rotated}
  \end{subfigure}
  \hfill
  \begin{subfigure}[b]{0.3\textwidth}
    \centering
    \includegraphics[width=\textwidth]{
      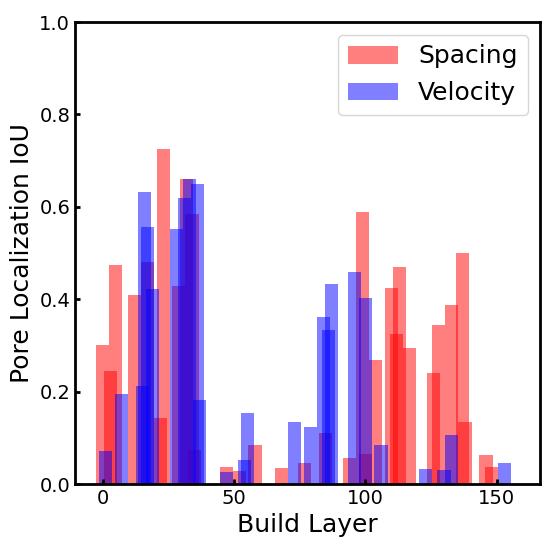
    }
    \caption{\textit{All}}
    \label{fig:localization_all_not_rotated}
  \end{subfigure}
  \\
  \begin{subfigure}[b]{0.3\textwidth}
    \centering
    \includegraphics[width=\textwidth]{
      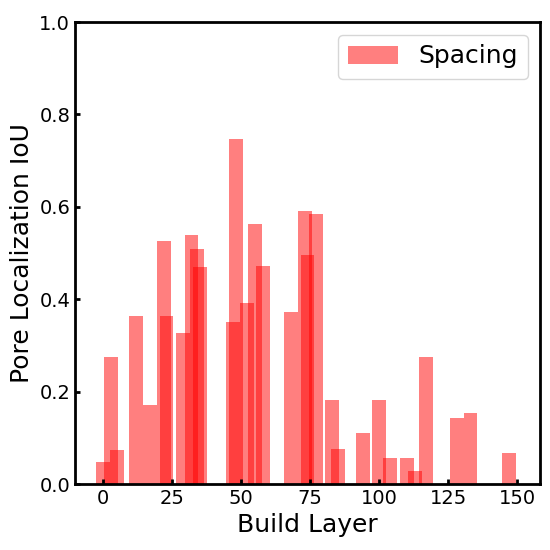
    }
    \caption{\textit{Spacing} (with transforms)}
    \label{fig:localization_spacing_rotated}
  \end{subfigure}
  \hfill
  \begin{subfigure}[b]{0.3\textwidth}
    \centering
    \includegraphics[width=\textwidth]{
      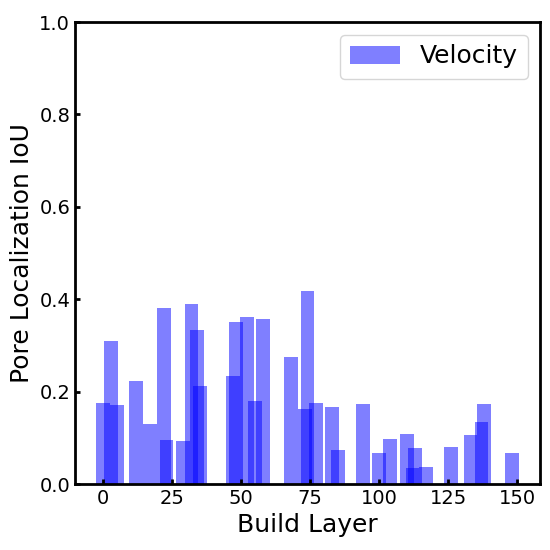
    }
    \caption{\textit{Velocity} (with transforms)}
    \label{fig:localization_velocity_rotated}
  \end{subfigure}
  \hfill
  \begin{subfigure}[b]{0.3\textwidth}
    \centering
    \includegraphics[width=\textwidth]{
      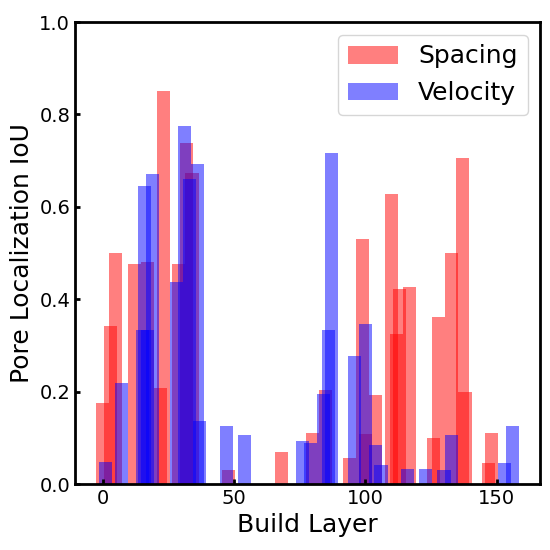
    }
    \caption{\textit{All} (with transforms)}
    \label{fig:localization_all_rotated}
  \end{subfigure}
  \hfill
  \caption{
    IoU trends by build layer for various datasets with and without rotational
    transforms.
  }
  \label{fig:pore_localization_results}
\end{figure}

\begin{table}
  \begin{tabular}{ll|ll|ll}
    \hline
    \rowcolor{gray!30}
                        &                   & \multicolumn{2}{|l|}{\textit{All Pores}}  & \multicolumn{2}{l}{\textit{$\mu + \sigma$ Pores}} \\
    \rowcolor{gray!30}
    Dataset             & Data Augmentation & Average IoU    & Max IoU                  & Average IoU    & Max IoU                          \\
    \hline                                 
    \textit{Spacing}    &  Rotational       & 0.25           & 0.75                     & 0.29           & \textbf{1.0}                     \\
                        &  None             & \textbf{0.28}  & 0.77                     & 0.28           & 1.0                              \\
    \rowcolor{gray!10}                     
    \textit{Velocity}   &  Rotational       & 0.16           & 0.42                     & 0.24           & 1.0                              \\
    \rowcolor{gray!10}                     
                        &  None             & 0.17           & 0.44                     & 0.29           & 1.0                              \\
    \textit{All}        &  Rotational       & 0.22           & \textbf{0.85}            & 0.32           & 1.0                              \\
                        &  None             & 0.21           & 0.72                     & \textbf{0.32}  & \textbf{1.0}                     \\
    \hline
  \end{tabular}
  \caption{
    Pore localization prediction performance metrics for ViViT Dense model
    trained on \textit{Spacing}, \textit{Velocity}, and \textit{All} sample
    datasets for 1000 epochs on both all segmented pores and pores with
    equivalent diameters greater than 1 standard deviation above the mean. The
    adjusted threshold to consider only pores with larger equivalent diameters
    improved prediction results in build layers with nominal process parameter
    and low resulting porosity.
  }
  \label{tab:pore_localization_vivit_dense_results}
\end{table}

\section{Conclusion and Future Work}
In this work we investigate the application of machine learning to
\textit{in-situ} thermal image process monitoring for the prediction of pore
count and pore localization. For this we utilized a CNN architecture and a
modified ViViT model with dense prediction heads for various dataset such as
\textit{Spacing}, \textit{Velocity}, and \textit{All}. For the task of pore
quantification, we have found that the \textit{Spacing} dataset provides the
greatest amount of signal within and models trained on the \textit{Velocity}
dataset produces the least amount of error. The pore localization task displayed
a similar trend with models trained on the \textit{Spacing} dataset achieving
the best overlap when evaluating \textit{All Pores}. The model trained on the
\textit{All} dataset showed better performance when evaluating on \textit{$\mu +
\sigma$ Pores}.

In both tasks, the effect of rotational transforms were minimal resulting in a
negligible difference in prediction outcomes. Our pore localization model
experienced improved performance with the application of a minimum pore ESD
threshold as it achieved higher average IoU scores, especially within areas of
the sample built with nominal processing parameters. These works show the
potential of utilizing \textit{in-situ} process monitoring techniques for faster
\textit{ex-situ} part certification and future work would aim to develop a more
robust digital twin achieving greater defect quantification and localization
precision over the entire sample.

\begin{acknowledgement}

Sandia National Laboratories is a multimission laboratory managed and operated
by National Technology \& Engineering Solutions of Sandia, LLC, a wholly owned
subsidiary of Honeywell International Inc., for the U.S. Department of Energy’s
National Nuclear Security Administration under contract
DE-NA0003525.
This paper describes objective technical results and analysis. Any subjective
views or opinions that might be expressed in the paper do not necessarily
represent the views of the U.S. Department of Energy or the United States
Government.

\end{acknowledgement}

\begin{suppinfo}

\section{Input Cropping and Label Downsampling}
The input thermal images were cropped from the original 85 px \texttimes\;60 px
shape down to a 64 px \texttimes\;64 px shape to fit the desired input shape of
the network. To achieve this 8 pixels (188.8 \textmu m) were cropped from both
the top and bottom of the image and 1 pixel (23.6 \textmu m) was removed from
the left hand side of the image. To match the cropped input, the corresponding
CT label was first cropped then downscaled to align with the expected 64 px
\texttimes\;64 px input. The crops of 8 pixels and 1 pixel were converted to
their voxel equivalents and rounded to 52 voxels and 7 voxels respectively. This
reduced the CT label a size of 520 voxels \texttimes\;423 voxels down to 416
voxels \texttimes\;416 voxels (Fig.
\ref{fig:input_cropping_label_downsampling}a). The resulting cropped CT was
downscaled by a factor of 24 down to 18 voxels \texttimes\;18 voxels along the x
and y directions and 1 voxel along the z direction. Further cropping was applied
to the x and y directions by 1 voxel on each side to reduce the label down to 16
voxels \texttimes\;16 voxels\texttimes\;1 voxel (Fig.
\ref{fig:input_cropping_label_downsampling}b).

\begin{figure}[bt]
  \centering
  \includegraphics[width=\textwidth]{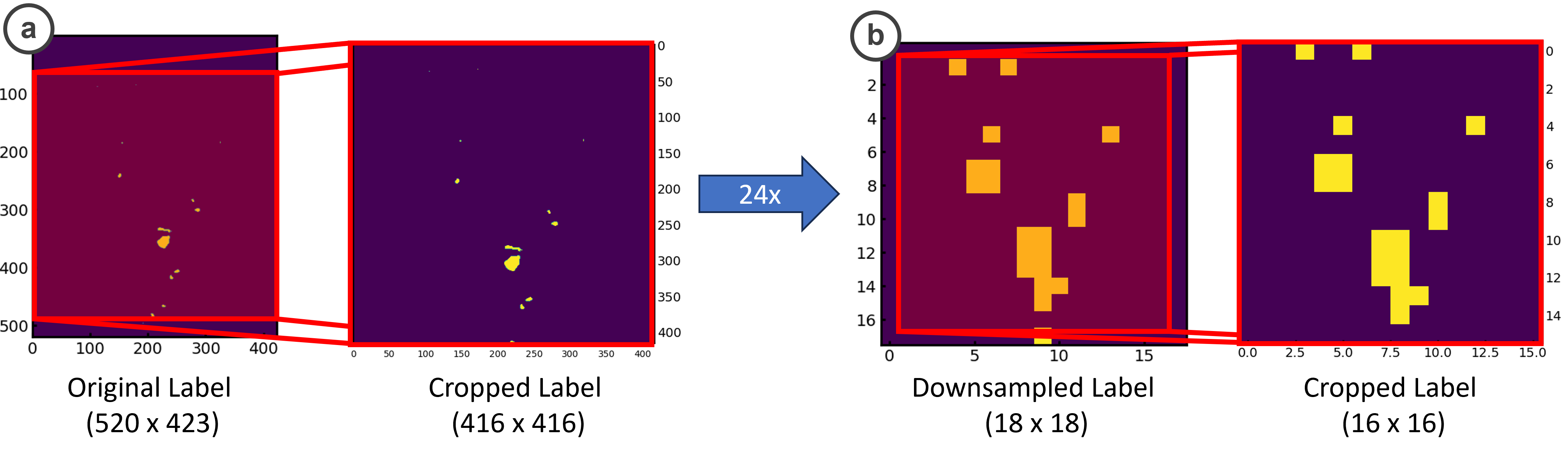}
  \caption{
  Relative porosity locations are derived from the raw CT data which are cropped
  to capture a consistent area to match the input thermal data (Fig.
  \ref{fig:input_cropping_label_downsampling}a). The CT data is further
  downsampled factor of 24 and cropped to provide a 16 x 16 label shape (Fig.
  \ref{fig:input_cropping_label_downsampling}b).
  }
  \label{fig:input_cropping_label_downsampling}
\end{figure}

\section{Calculating absolute temperature estimations with data
obtained through Stratonics pyrometer}

The estimation of absolute temperature was achieved with grey-body assumption
where the emissivity remains constant at various wavelengths. Thus, with proper
calibration using a NIST-traceable tungsten lamp, the pyrometer was able to read
out a temperature estimation within 4\% accuracy for stainless steel.
\cite{mitchell_linking_2020}. This temperature is estimated with the following
formula for hybrid mode temperature estimation (Eq. \ref{eq:t_h}) where $p_2$ is
a constant held at 14388 nm-K. Wien's approximation \cite{ikeuchi_computer_2021}
of Planck's law is used to evaluate $A_\lambda$ (Eq. \ref{eq:a_lambda}) which
takes into consideration emissivity and the instrument's detection factor with
$\bar{I_1}$ representing the average intensity calculated over region
$\Omega_p$. $I_1$ and $I_2$ here represent the the radiance from images over
$\Omega_p$, for $\lambda_1 = 750 \text{nm}$ and $\lambda_2 = 900 \text{nm}$
respectively. The temperature ratio $T_R$ (Eq. \ref{eq:t_r}) is calculated with
$R$, the average of radiance from images $I_1$ and $I_2$ over $\Omega_p$, and
hardware constants for contour levels $c_1$ and $c_2$ obtained from least
squares fitting over calibration data\cite{zalameda_four-color_2016}. Contour
level defines the region $\Omega_p$ with a $\beta$ value between 0 and 1 through
means of marching squares (Eq. \ref{eq:c}). Our dataset utilizes images taken at
$\beta = 0.7$ for its most accurate approximation of the meltpool by single
contour as opposed to the multiple contours that appear at $\beta = 0.3$.

\begin{equation}
  T_H = \frac{p_2}{\lambda_2\ln\left({A_\lambda/I_2}\right)}
  \label{eq:t_h}
\end{equation}
\begin{equation}
  A_\lambda =  \bar{I}_1 e^{p_2/\lambda_1T_R}
  \label{eq:a_lambda}
\end{equation}
\begin{equation}
  T_R = \frac{1}{c_1 \ln R + c_2}
  \label{eq:t_r}
\end{equation}
\begin{equation}
  c = \beta \times \text{max}\left(I\right)
  \label{eq:c}
\end{equation}

\subsection{
  Sample alignment of thermal images to CT data along the X and Y
  axes
}
CT data for both samples had a shape of 1410 \texttimes\;900 \texttimes\;430
voxels with values ranging from 0 - 255 depending on data selection type of
pore, sample, or pore segmented sample. Common alignment between the
\textit{in-situ} pyrometry data and \textit{ex-situ} CT data was established by
converting the pyrometry data into voxels and applying a consistent offset.
The \textit{in-situ} pyrometry data was converted from pixels (80 px
\texttimes\;65 px) to microns (1680 \textmu m \texttimes\;1365 \textmu m) to
voxels (462.81 voxels \texttimes\;376.03 voxels). The CT data was then shifted
to align with the pyrometry data by -18 voxels \texttimes\;0 voxels
\texttimes\;63 voxels for the \textit{Spacing} dataset and -14 voxels
\texttimes\;-2 voxels \texttimes\;73 voxels for the \textit{Velocity} dataset.

\subsection{
  IoU scores of models trained on equivalent sphere diameter pores 1 standard
  deviation above the mean (\textit{$\mu + \sigma$ Pores})
}
\begin{figure}[bt]
  \centering
  \begin{subfigure}[b]{0.3\textwidth}
    \centering
    \caption{\textit{Spacing}}
    \includegraphics[width=\textwidth]{
      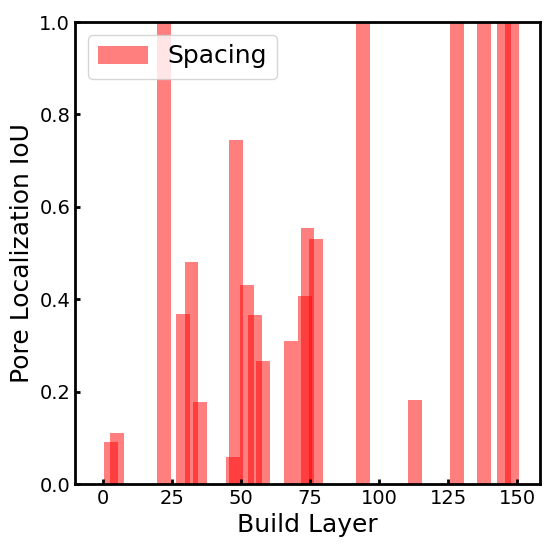
    }
    \label{fig:localization_spacing_not_rotated_thresholded}
  \end{subfigure}
  \hfill
  \begin{subfigure}[b]{0.3\textwidth}
    \centering
    \caption{\textit{Velocity}}
    \includegraphics[width=\textwidth]{
      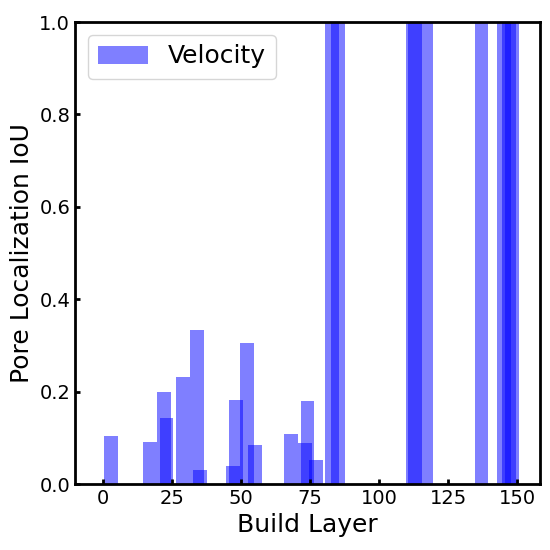
    }
    \label{fig:localization_velocity_not_rotated_thresholded}
  \end{subfigure}
  \hfill
  \begin{subfigure}[b]{0.3\textwidth}
    \centering
    \caption{\textit{All}}
    \includegraphics[width=\textwidth]{
      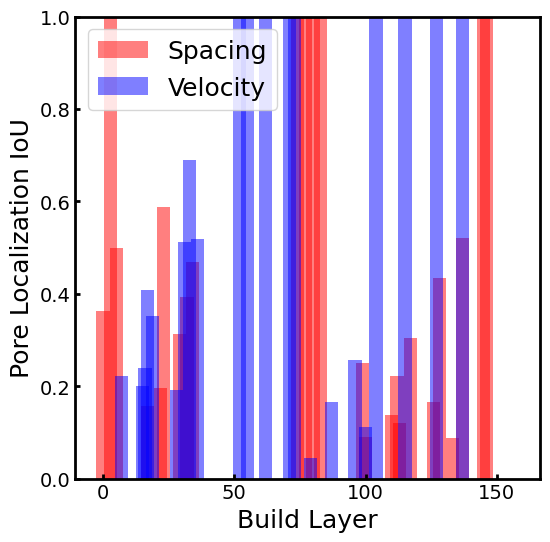
    }
    \label{fig:localization_all_not_rotated_thresholded}
  \end{subfigure}
  \\
  \begin{subfigure}[b]{0.3\textwidth}
    \centering
    \includegraphics[width=\textwidth]{
      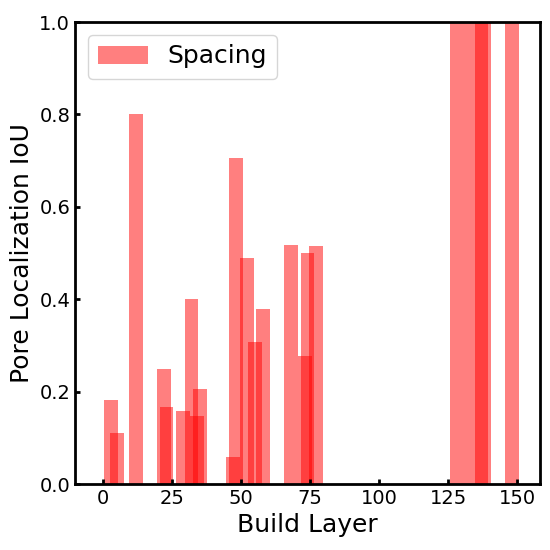
    }
    \caption{\textit{Spacing} (with transforms)}
    \label{fig:localization_spacing_rotated_thresholded}
  \end{subfigure}
  \hfill
  \begin{subfigure}[b]{0.3\textwidth}
    \centering
    \includegraphics[width=\textwidth]{
      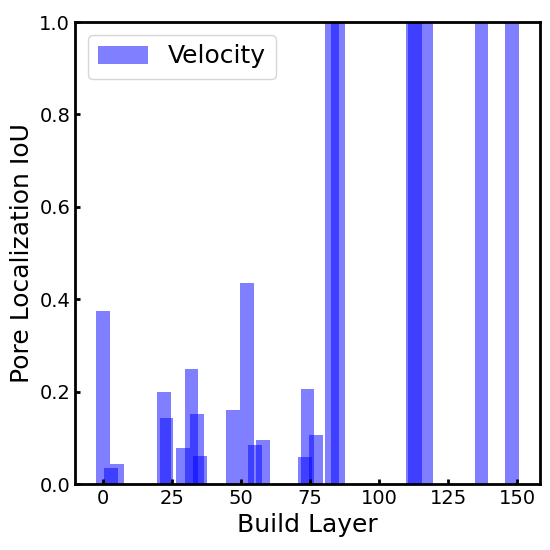
    }
    \caption{\textit{Velocity} (with transforms)}
    \label{fig:localization_velocity_rotated_thresholded}
  \end{subfigure}
  \hfill
  \begin{subfigure}[b]{0.3\textwidth}
    \centering
    \includegraphics[width=\textwidth]{
      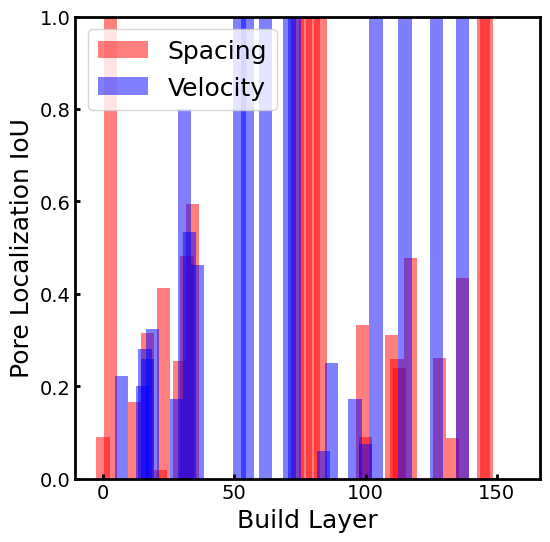
    }
    \caption{\textit{All} (with transforms)}
    \label{fig:localization_all_rotated_thresholded}
  \end{subfigure}
  \caption{
    IoU trends by build layer for various datasets with and without rotational
    transforms with the application of minimum equivalent diameter pore
    threshold.
  }
  \label{fig:pore_localization_results_thresholded}
\end{figure}

\end{suppinfo}

\bibliography{references}

\end{document}